\documentclass[letterpaper]{article} % DO NOT CHANGE THIS
\usepackage{aaai2026}  % DO NOT CHANGE THIS
\usepackage{times}  % DO NOT CHANGE THIS
\usepackage{helvet}  % DO NOT CHANGE THIS
\usepackage{courier}  % DO NOT CHANGE THIS
\usepackage[hyphens]{url}  % DO NOT CHANGE THIS
\usepackage{graphicx} % DO NOT CHANGE THIS
\usepackage{natbib}  % DO NOT CHANGE THIS AND DO NOT ADD ANY OPTIONS TO IT
\usepackage{caption} % DO NOT CHANGE THIS AND DO NOT ADD ANY OPTIONS TO IT
\usepackage{amsmath}
\DeclareMathOperator*{\argmax}{arg\,max}
\usepackage{algorithm}
\usepackage{algorithmic}
\usepackage{colortbl}
\usepackage{multirow}

\usepackage{newfloat}
\usepackage{listings}
\DeclareCaptionStyle{ruled}{labelfont=normalfont,labelsep=colon,strut=off} % DO NOT CHANGE THIS
\floatstyle{ruled}
\newfloat{listing}{tb}{lst}{}
\floatname{listing}{Listing}
\definecolor{reasonrow}{RGB}{210,230,250}

\title{Reasoning Is All You Need for Urban Planning AI}

\author{
    Sijie Yang\textsuperscript{\rm 1},
    Jiatong Li\textsuperscript{\rm 1,2},
    Filip Biljecki\textsuperscript{\rm 1,3,*}
}
\affiliations{
    \textsuperscript{\rm 1}Department of Architecture, National University of Singapore\\
    \textsuperscript{\rm 2}School of Architecture, Tsinghua University\\
    \textsuperscript{\rm 3}Department of Real Estate, National University of Singapore\\
    \textsuperscript{*}Corresponding author: filip@nus.edu.sg\\
}

\usepackage{bibentry}
\begin{document}

\nocopyright  % Remove copyright notice for arXiv version
\maketitle

\begin{abstract}
AI has proven highly successful at urban planning \textit{analysis}---learning patterns from data to predict future conditions. The next frontier is AI-assisted \textit{decision-making}: agents that recommend sites, allocate resources, and evaluate trade-offs while reasoning transparently about constraints and stakeholder values. Recent breakthroughs in reasoning AI---CoT\footnote{Chain-of-Thought: A prompting technique that encourages large language models to break down complex reasoning into intermediate steps \citep{wei2023chainofthought}} prompting, ReAct\footnote{Reasoning and Acting: A framework that interleaves reasoning traces with task-specific actions in language models \citep{yao2023react}}, and multi-agent collaboration frameworks---now make this vision achievable.

This position paper presents the Agentic Urban Planning AI Framework for reasoning-capable planning agents that integrates three cognitive layers (Perception, Foundation, Reasoning) with six logic components (Analysis, Generation, Verification, Evaluation, Collaboration, Decision) through a multi-agents collaboration framework. We demonstrate why planning decisions require explicit reasoning capabilities that are value-based (applying normative principles), rule-grounded (guaranteeing constraint satisfaction), and explainable (generating transparent justifications)---requirements that statistical learning alone cannot fulfill. We compare reasoning agents with statistical learning, present a comprehensive architecture with benchmark evaluation metrics, and outline critical research challenges. This framework shows how AI agents can augment human planners by systematically exploring solution spaces, verifying regulatory compliance, and deliberating over trade-offs transparently---not replacing human judgment but amplifying it with computational reasoning capabilities.
\end{abstract}


% Direct opening - no "Introduction" heading
By 2050, 68\% of humanity will live in cities \citep{un2019world}. AI has transformed urban planning analytics \citep{jha2021review, sanchez2023prospects, peng2024pathway, wang2025generative}, achieving unprecedented accuracy in prediction tasks. Yet a critical question emerges as AI capabilities advance: \textit{Can statistical learning alone support planning decisions, or do we need explicit reasoning capabilities?} We argue that planning decisions demand reasoning agents that are \textbf{value-based}, \textbf{rule-grounded}, and \textbf{explainable}---capabilities that statistical pattern learning alone cannot provide.

\textbf{Our contributions}: (1) We compare reasoning agents with statistical learning, demonstrating why explicit reasoning is foundational for planning decisions; (2) We present the \textit{Agentic Urban Planning AI Framework}---a three-layer cognitive architecture (Perception, Foundation, Reasoning) integrating six logic components (Analysis, Generation, Verification, Evaluation, Collaboration, Decision) through a multi-agents collaboration framework, formalised with algorithms and evaluation metrics; (3) We outline five critical research challenges and a path forward for building reasoning-capable planning agents that augment human judgment with computational reasoning capabilities.

\section{AI for UP: From Analytics to Decision Support}

AI's success in urban planning analytics is undeniable---predicting traffic with RNNs\footnote{Recurrent Neural Networks} \citep{lv2018lcrnn}, classifying land uses with CNNs\footnote{Convolutional Neural Networks} \citep{zhang2018objectbased}, forecasting building carbon emissions in cities with GNNs\footnote{Graph Neural Networks} \citep{yap2025revealing}, assessing urban heat islands \citep{xu2022comparing}, evaluating thermal comfort \citep{yang2025thermal}, analysing urban morphology and street environment impacts \citep{qiu2022subjective,yang2023role}, identifying street activity patterns \citep{li2025identifying}, and developing liveability indices \citep{lei2025developing}. These \textit{pattern-based, data-driven} GeoAI approaches \citep{liu2022review} excel at learning complex correlations from historical urban data---predicting \textit{what will happen}.

The frontier now is AI-assisted \textit{decision-making}: recommending sites, allocating resources, evaluating trade-offs \citep{zhu2025plangpt,liu2025urban}. Recent breakthroughs in reasoning AI make this achievable. CoT prompting \citep{wei2023chainofthought} generates intermediate reasoning steps; ReAct \citep{yao2023react} interleaves reasoning with tool-augmented actions; multi-agent collaboration frameworks like AutoGen \citep{wu2024autogen} enable coordinated deliberation among specialized agents. These techniques enable AI agents to deliberate transparently, guarantee regulatory compliance, and generate explainable justifications---capabilities planning decisions demand.

\begin{figure}[htbp]
    \centering
    \includegraphics[width=0.95\columnwidth]{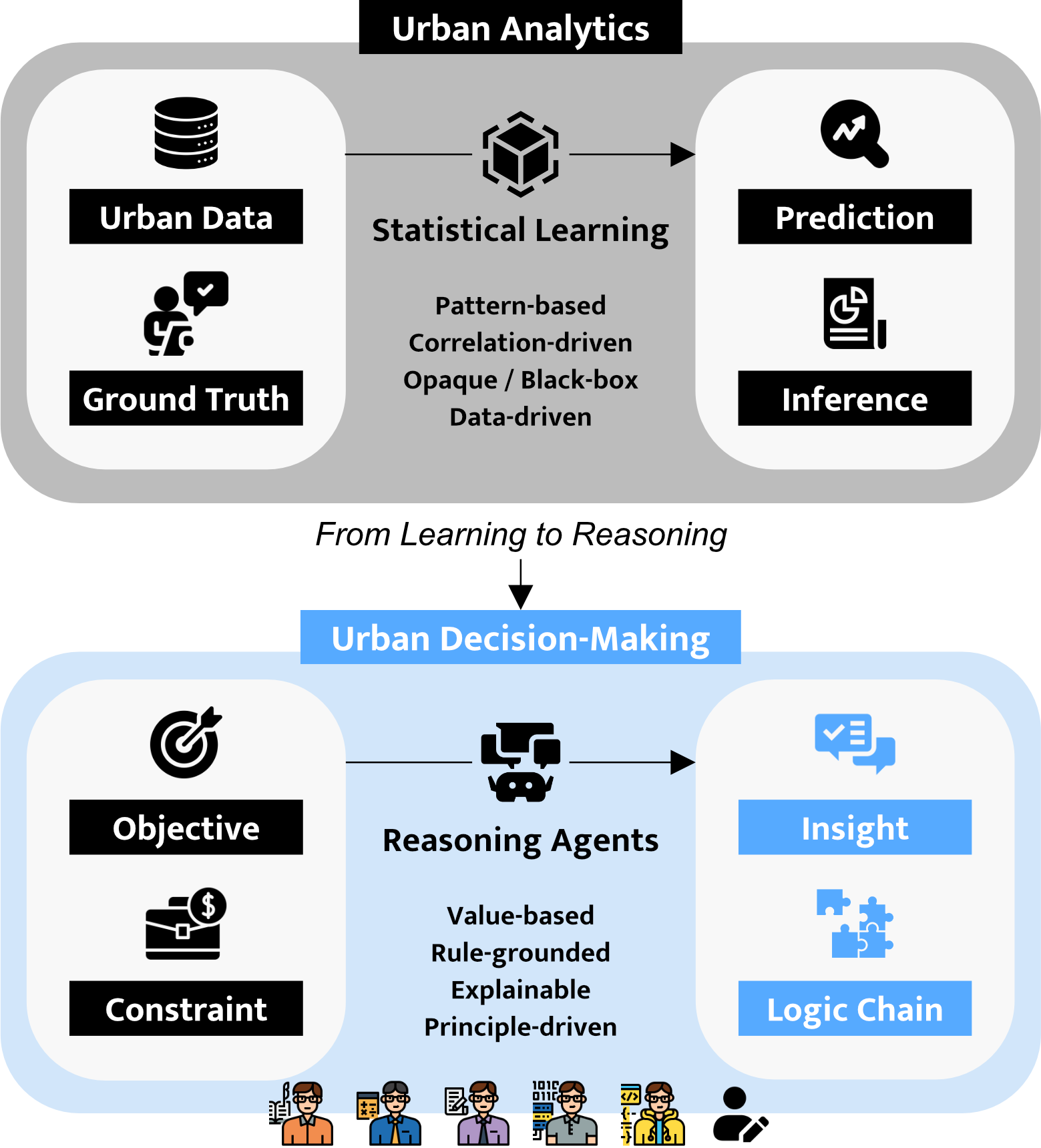}
    \caption{AI's dual role in urban planning: analysis (prediction tasks) and decision support (recommendation tasks with explicit reasoning).}
    \label{fig:paradigm-flow}
\end{figure}

We distinguish two paradigms by \textit{how decisions are made} (Figure \ref{fig:paradigm-flow}): \textbf{Statistical learning} refers to systems which learn decision patterns from historical data through statistical correlations. These \textit{pattern-based, data-driven} approaches excel at prediction but have opaque decision processes. They can replicate historical allocations, detect likely violations, and recommend solutions---but struggle to apply normative principles, guarantee constraint satisfaction, or explain reasoning chains. \textbf{Reasoning Agents} refer to systems which generate explicit reasoning traces to reach decisions, using LLM-based reasoning (CoT, ReAct). These agents can challenge unjust patterns, resolve contradictory rules, and explain counterfactual logic---capabilities statistical learning lacks.

\begin{table}[!htbp]
    \centering
    \renewcommand{\arraystretch}{1.4}
    \begin{small}
    \setlength{\tabcolsep}{5pt}
    \begin{tabular}{p{4.8cm}!{\color{gray!60}\vrule}cc}
    \hline
    \textbf{Planning Decision Task} & \textbf{Statistical} & \textbf{Reasoning} \\
     & \textbf{Learning} & \textbf{Agents} \\
    \hline\hline
    \textbf{Value-Based \& Principle-Driven} & & \\
    \textit{Equity-driven resource allocation} & & \\
    \quad Replicate historical allocation & $\bullet$ & $\bullet$ \\
    \quad Apply equity principles & $\circ$ & $\bullet$ \\
    \quad Challenge unjust patterns & -- & $\bullet$ \\
    \textit{Novel urban planning context} & & \\
    \quad Transfer similar patterns & $\bullet$ & $\bullet$ \\
    \quad Reason from first principles & -- & $\bullet$ \\
    \textit{Competing value prioritisation} & & \\
    \quad Learn stakeholder preferences & $\bullet$ & $\bullet$ \\
    \quad Deliberate on normative priorities & -- & $\bullet$ \\
    \hline\hline
    \textbf{Rule-Grounded} & & \\
    \textit{Zoning regulation compliance} & & \\
    \quad Detect likely violations & $\bullet$ & $\bullet$ \\
    \quad Guarantee zero violations & -- & $\bullet$ \\
    \textit{Multi-constraint optimization} & & \\
    \quad Find feasible solutions & $\bullet$ & $\bullet$ \\
    \quad Verify all constraints satisfied & $\circ$ & $\bullet$ \\
    \textit{Contradictory rule resolution} & & \\
    \quad Flag conflicting requirements & $\circ$ & $\bullet$ \\
    \quad Resolve using legal reasoning & -- & $\bullet$ \\
    \hline\hline
    \textbf{Explainable} & & \\
    \textit{Decision justification to public} & & \\
    \quad Provide recommendations & $\bullet$ & $\bullet$ \\
    \quad Generate readable rationale & $\circ$ & $\bullet$ \\
    \textit{Causal impact chain explanation} & & \\
    \quad Predict outcomes & $\bullet$ & $\bullet$ \\
    \quad Trace cause-effect reasoning & -- & $\bullet$ \\
    \textit{``What-if'' scenario analysis} & & \\
    \quad Simulate alternative outcomes & $\bullet$ & $\bullet$ \\
    \quad Explain counterfactual logic & -- & $\bullet$ \\
    \hline\hline
    \end{tabular}
    \end{small}
    \caption{Paradigm comparison on planning decision tasks. Legend: $\bullet$ = well-supported; $\circ$ = limited; -- = not supported. Both paradigms can address planning tasks, but reasoning agents provide value-based deliberation, rule-grounded verification, and explainable justification.}
    \label{tab:paradigm-comparison}
\end{table}

Why do planning decisions specifically require reasoning agents? Table \ref{tab:paradigm-comparison} compares both paradigms across nine decision tasks. While statistical learning handles many tasks effectively, reasoning agents provide three critical capabilities planning demands:

\textbf{Value-Based \& Principle-Driven}: Planning decisions are \textit{normative}---reflecting values, principles, and long-term visions rather than learned patterns. Consider equity-driven resource allocation: statistical learning replicates historical allocations, but reasoning agents apply equity principles and challenge unjust patterns embedded in historical data. For novel contexts without precedent---climate adaptation, emerging technologies---reasoning agents apply first principles when historical patterns provide insufficient guidance. When competing values conflict, reasoning agents deliberate on normative priorities rather than merely learning stakeholder preferences from past data.

\textbf{Rule-Grounded}: Planning operates under hard constraints that must be satisfied with certainty. For zoning regulation compliance, statistical learning detects likely violations, but reasoning agents guarantee zero violations through formal verification. Multi-constraint optimization requires verifying that \textit{all} constraints are satisfied simultaneously, not just finding feasible solutions. When contradictory rules arise, reasoning agents resolve conflicts using legal reasoning rather than flagging inconsistencies without resolution.

\textbf{Explainable}: Planning decisions require transparent justifications for legal review and public scrutiny. For decision justification to the public, reasoning agents generate readable rationales explaining \textit{why} recommendations were made. Causal impact chain explanation demands tracing cause-effect reasoning, not just predicting outcomes. ``What-if'' scenario analysis requires explaining counterfactual logic to support deliberative decision processes. The Sidewalk Labs Toronto smart city proposal's opacity challenges underscore the necessity of explainable AI in planning contexts.

\section{Reasoning-Capable Urban Planning Agents}

We now present a comprehensive architecture that integrates contemporary reasoning techniques with symbolic constraint solving to enable transparent, verifiable, and collaborative planning decision support. As illustrated in Figure \ref{fig:architecture}, our framework comprises three cognitive layers and six logic components that operate through human-AI collaborative workflows (Figure \ref{fig:collaboration}).

\subsection{Agentic Urban Planning AI Framework}

\begin{table*}[!htbp]
    \centering
    \renewcommand{\arraystretch}{1.3}
    \begin{small}
    \setlength{\tabcolsep}{4pt}
    \begin{tabular}{>{\centering\arraybackslash}p{1.8cm}>{\centering\arraybackslash}p{3.2cm}!{\color{gray!40}\vrule}>{\centering\arraybackslash}p{4.9cm}!{\color{gray!40}\vrule}>{\centering\arraybackslash}p{4.5cm}!{\color{gray!60}\vrule}>{\centering\arraybackslash}p{2cm}}
    \hline
    \textbf{Cognitive Layer} & \textbf{Sub-Module} & \textbf{Representative AI} & \textbf{Urban Planning Function} & \textbf{Planning Stage} \\
    \hline\hline
    \multirow{6}{*}{\parbox{1.8cm}{\centering\textbf{Perception Layer}}} 
    & \multirow{2}{*}{\parbox{3.2cm}{\centering Visual Perception}} 
    & SAM \citep{kirillov2023segment} 
    & Urban imagery segmentation 
    & \multirow{6}{*}{\parbox{2cm}{\centering Data Collection}} \\
    & & ViT \citep{dosovitskiy2020image} & Urban imagery embedding & \\
    \cline{2-4}
    & \multirow{2}{*}{\parbox{3.2cm}{\centering Cross-Modal Fusion}} 
    & CLIP \citep{radford2021learning} & Vision-language alignment & \\
    & & BLIP-2 \citep{li2023blip2} & Multi-modal understanding & \\
    \cline{2-4}
    & \multirow{2}{*}{\parbox{3.2cm}{\centering 3D Reconstruction}} 
    & NeRF \citep{mildenhall2020nerf} & 3D urban scene reconstruction & \\
    & & 3DGS \citep{kerbl20233d} & Urban geometry modeling & \\
    \hline\hline
    \multirow{8}{*}{\parbox{1.8cm}{\centering\textbf{Foundation Layer}}} 
    & \multirow{2}{*}{\parbox{3.2cm}{\centering Statistical Learning}} 
    & XGBoost \citep{chen2016xgboost} 
    & Tabular data prediction 
    & \multirow{8}{*}{\parbox{2cm}{\centering Knowledge}} \\
    & & SHAP \citep{lundberg2017unified} & Model interpretation & \\
    \cline{2-4}
    & \multirow{2}{*}{\parbox{3.2cm}{\centering Large Language Models}} 
    & Qwen 3 \citep{yang2025qwen3} 
    & Planning document analysis & \\
    & & Llama 3 \citep{grattafiori2024llama} & Policy semantic understanding & \\
    \cline{2-4}
    & \multirow{2}{*}{\parbox{3.2cm}{\centering RAG}} 
    & RAG \citep{lewis2020retrievalaugmented} 
    & Knowledge retrieval & \\
    & & LangChain \citep{chase2022langchain} & Regulation query system & \\
    \cline{2-4}
    & \multirow{2}{*}{\parbox{3.2cm}{\centering Simulation \& RL}}
    & DQN \citep{mnih2015humanlevel} 
    & Policy learning & \\
    & & PPO \citep{schulman2017proximal} & Multi-objective optimization & \\
    \hline\hline
    \multirow{12}{*}{\parbox{1.8cm}{\centering\textbf{Reasoning Layer}}} 
    & \multirow{2}{*}{\parbox{3.2cm}{\centering Cognitive Reasoning}} 
    & CoT \citep{wei2023chainofthought} 
    & Step-by-step reasoning 
    & \multirow{2}{*}{\parbox{2cm}{\centering All Stages}} \\
    &  & ToT \citep{yao2023treea} & Alternative exploration & \\
    \cline{2-5}
    & \multirow{2}{*}{\parbox{3.2cm}{\centering Goal-Oriented Planning}} 
    & LATS \citep{zhou2024language} 
    & Autonomous task planning 
    & \multirow{2}{*}{\parbox{2cm}{\centering Analysis, Generation}} \\
    &  & Voyager \citep{wang2023voyager} & Embodied learning & \\
    \cline{2-5}
    & \multirow{2}{*}{\parbox{3.2cm}{\centering Tool-Augmented}} 
    & ReAct \citep{yao2023react} 
    & External tool invocation 
    & \multirow{2}{*}{\parbox{2cm}{\centering Generation, Verification}} \\
    &  & Toolformer \citep{schick2023toolformer} & API-based computation & \\
    \cline{2-5}
    & \multirow{2}{*}{\parbox{3.2cm}{\centering Normative Reasoning}} 
    & Const AI \citep{bai2022constitutional} 
    & Value-aligned reasoning 
    & \multirow{2}{*}{\parbox{2cm}{\centering Evaluation, Decision}} \\
    &  & RLHF \citep{christiano2017deep} & Preference learning & \\
    \cline{2-5}
    & \multirow{2}{*}{\parbox{3.2cm}{\centering Multi-Agent System}} 
    & AutoGen \citep{wu2023autogen} 
    & Multi-agent collaboration 
    & \multirow{2}{*}{\parbox{2cm}{\centering Collaboration, Decision}} \\
    &  & MetaGPT \citep{hong2023metagpt} & Role-based coordination & \\
    \cline{2-5}
    & \multirow{2}{*}{\parbox{3.2cm}{\centering Test-Time Reasoning}} 
    & GPT-o1 \citep{openai2025o1} 
    & Advanced reasoning 
    & \multirow{2}{*}{\parbox{2cm}{\centering Verification, Decision}} \\
    &  & DeepSeek-R1 \citep{guo2025deepseekr1} & RL-based reasoning & \\
    \hline\hline
    \end{tabular}
    \end{small}
    \caption{Agentic urban planning AI framework: A three-layer cognitive architecture for planning. The framework progresses through \textbf{Perception} (data collection), \textbf{Foundation} (knowledge building with statistical learning, LLMs, RAG, and simulation), and \textbf{Reasoning} (agentic AI for urban planning tasks). The reasoning layer comprises six functional stages: analysis, generation, verification, evaluation, collaboration, and decision. Representative AI models are mapped to their corresponding urban planning functions and stages. }
    \label{tab:ai-categories}
\end{table*}

Table \ref{tab:ai-categories} presents a comprehensive agentic urban planning AI framework organised as a three-layer cognitive architecture: \textbf{Perception $\rightarrow$ Foundation $\rightarrow$ Reasoning (Agentic AI)}. This framework applies urban planning reasoning with agentic AI---LLM\footnote{Large Language Model}-based autonomous systems capable of perception-grounded data representation, external tool-augmented analysis, and value-aligned decision-making. While perception and foundation layers capture and organise urban knowledge, the reasoning layer embodies goal-directed agents that deliberate, verify, and act upon urban problems. Critically, we position reinforcement learning dually: in the Foundation layer as environment modelling for simulation and behaviour learning, and in the Reasoning layer as policy optimisation for strategic decision-making. Similarly, RAG serves as the ``memory interface'' of the Foundation layer, providing retrievable urban knowledge to agentic systems, while LLMs constitute the core reasoning engine built upon foundational pretraining. This architecture clarifies that Agentic AI = LLM (Core) + RAG (Memory) + Tools (Action) + RL (Feedback) + Values (Constraint).

\begin{figure}[htbp]
\centering
\includegraphics[width=0.95\columnwidth]{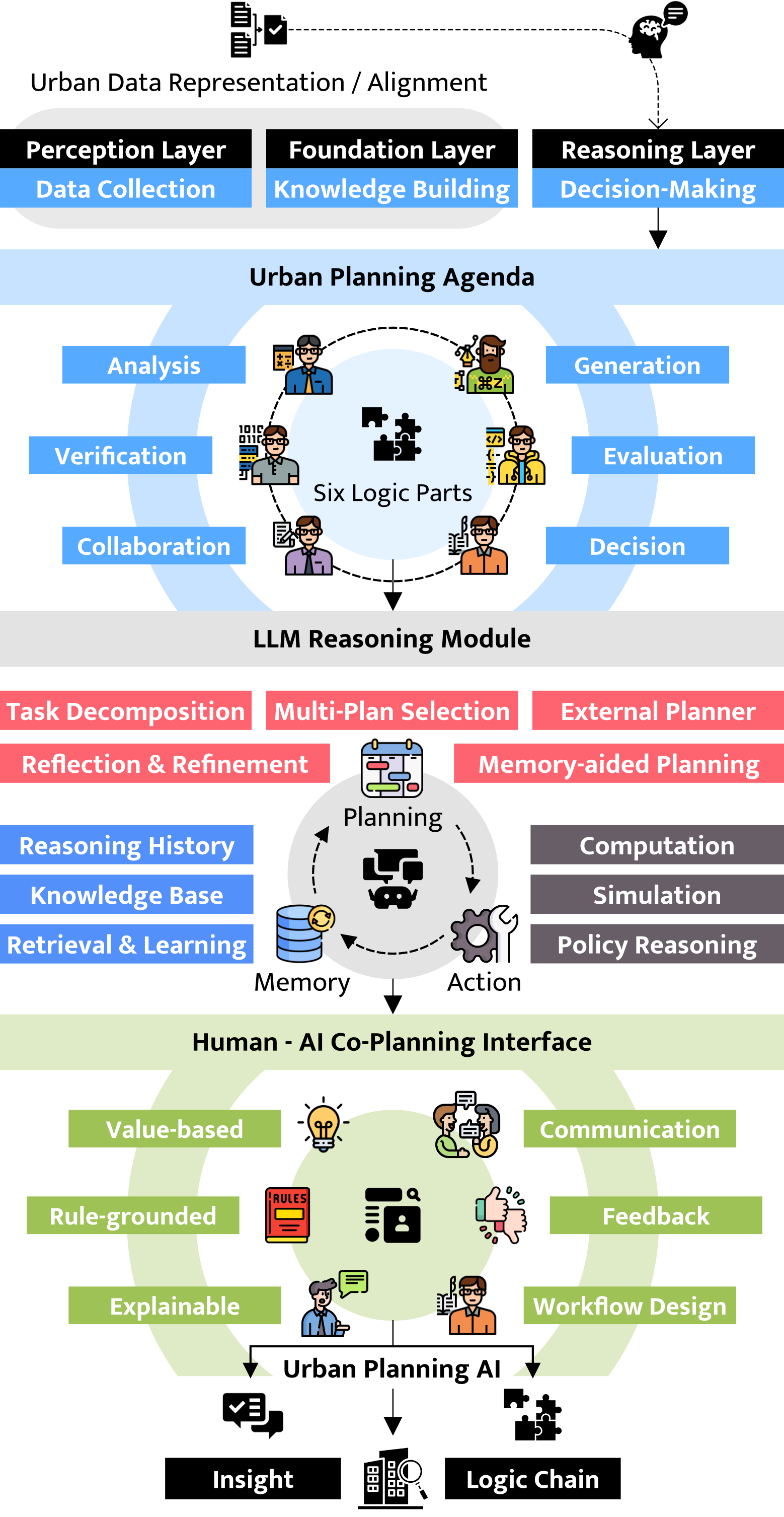}
\caption{Agentic urban planning AI framework for reasoning-capable urban planning. The architecture comprises three cognitive layers (Perception, Foundation, Reasoning) and six logic components (Analysis, Generation, Verification, Evaluation, Collaboration, Decision) integrated through a human-AI co-planning interface supporting value-based, rule-grounded, and explainable decision-making.}
\label{fig:architecture}
\end{figure}

Beyond these three cognitive layers, the framework (illustrated in Figure \ref{fig:architecture}) integrates six specialized logic components that orchestrate planning deliberation.

\textbf{Three Cognitive Layers}: The framework progresses through three complementary layers aligned with urban planning stages:

\textit{Perception Layer (Data Collection)}: This foundation layer collects and processes multi-modal urban data \citep{yang2025urban}. Computer vision models (SAM, ViT) extract spatial information from satellite imagery and street-level photos \citep{liang2025openfacades}. Multi-modal models (CLIP, BLIP-2) link visual data with textual descriptions. 3D reconstruction techniques (NeRF, 3DGS) create detailed spatial representations. This layer transforms raw urban data into structured, machine-interpretable formats that inform downstream reasoning.

\textit{Foundation Layer (Knowledge Building)}: Statistical learning models build predictive knowledge from historical data. XGBoost and SHAP provide interpretable predictions and feature importance for traffic patterns, development impacts, and cost estimation. Large language models (Qwen, Llama 3) synthesize planning knowledge from regulatory documents, guidelines, and precedents \citep{hou2025urban, zheng2025urban}. Recent unified multimodal models \citep{xie2024showo,zhou2024transfusion} enable integrated understanding and generation across modalities. RAG and LangChain enable retrieval of relevant planning knowledge to support reasoning. This layer constructs the knowledge base that reasoning agents query during decision-making.

\textit{Reasoning Layer (Decision-Making)}: This layer performs explicit logical reasoning for planning decisions. Cognitive reasoning agents (DQN, PPO) learn strategic decision policies. Chain-of-Thought and Tree-of-Thought reasoning decompose complex planning problems into verifiable steps. Tool-using agents (ReAct, Toolformer) invoke constraint solvers and simulation tools. Multi-agent systems (AutoGen, MetaGPT) coordinate specialized reasoning agents \citep{wang2025cartoagent}. Value-aligned reasoning (Constitutional AI, RLHF) ensures decisions reflect planning principles. Advanced reasoning models (GPT-o1, DeepSeek-R1) provide process supervision for reasoning quality.

\textbf{Six Logic Components}: Building on these three layers, the reasoning architecture orchestrates six specialized components that correspond to distinct planning deliberation stages (see Figure \ref{fig:architecture} and Table \ref{tab:paradigm-comparison}): \textit{Analysis} conducts spatial and multi-criteria analysis; \textit{Generation} produces planning alternatives through constrained search; \textit{Verification} formally verifies regulatory compliance using symbolic solvers; \textit{Evaluation} assesses proposals against normative criteria (sustainability, equity, resilience); \textit{Collaboration} facilitates multi-stakeholder dialogue and consensus-building \citep{qian2023ai}; and \textit{Decision} synthesizes reasoning chains into actionable recommendations with explicit trade-offs. These components operate iteratively, enabling planners to critique reasoning, adjust priorities, and refine proposals through bidirectional human-AI interaction.

\textbf{Formal Problem Definition}. We formalize the urban planning decision problem as a constrained multi-objective optimization with explicit reasoning requirements. Given:
\begin{itemize}
\item A planning context $\mathcal{C} = \langle \mathcal{D}, \mathcal{K}, \mathcal{S} \rangle$ comprising spatial data $\mathcal{D}$, planning knowledge $\mathcal{K}$, and stakeholder input $\mathcal{S}$
\item A set of hard constraints $\mathcal{H} = \{h_1, h_2, \ldots, h_m\}$ representing regulatory requirements (zoning codes, environmental standards)
\item A set of soft objectives $\mathcal{O} = \{o_1, o_2, \ldots, o_n\}$ capturing normative criteria (equity, sustainability, liveability)
\end{itemize}

The reasoning-capable planning agent seeks to generate a proposal $p \in \mathcal{P}$ along with an explicit reasoning chain $r \in \mathcal{R}$ such that:

\begin{equation}
\begin{split}
p^*, r^* = \argmax_{p \in \mathcal{P}, r \in \mathcal{R}} \left[ \sum_{i=1}^{n} w_i \cdot o_i(p) \right] \\
\text{subject to} \quad \forall h_j \in \mathcal{H}: h_j(p) = \text{True}
\end{split}
\end{equation}

where $w_i$ are stakeholder-specified objective weights, and crucially, the reasoning chain $r$ must satisfy:
\begin{equation}
\text{Valid}(r) \land \text{Complete}(r) \land \text{Traceable}(r, p, \mathcal{C})
\end{equation}

This formulation distinguishes reasoning agents from statistical learning: reasoning chains $r$ provide explicit, verifiable justifications linking context $\mathcal{C}$ to proposal $p$, enabling human inspection and critique.

\subsection{Multi-Agents Collaboration Framework}

\begin{figure*}[!htbp]
    \centering
    \includegraphics[width=0.95\textwidth]{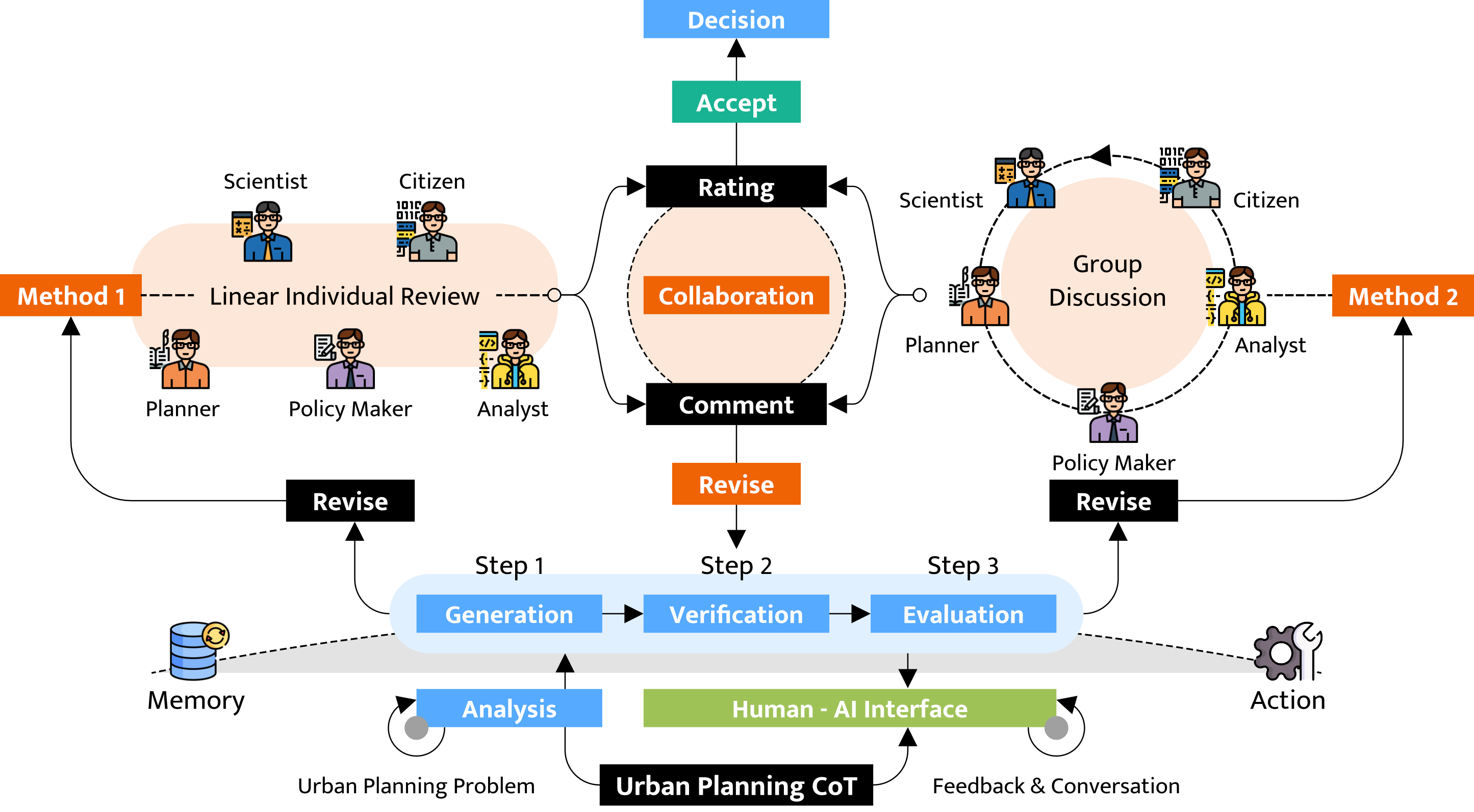}
    \caption{Multi-agents collaboration framework implementing the Collaboration component of the agentic urban planning AI framework. The framework supports two collaboration methods: linear individual review (Method 1) and group discussion (Method 2). Six logic parts (Analysis, Generation, Verification, Evaluation, Collaboration, Decision) operate through the human-AI interface, enabling iterative refinement via rating, commenting, and revision across three cognitive layers.}
    \label{fig:collaboration}
\end{figure*}

As illustrated in Figure \ref{fig:collaboration}, the Collaboration component of the Agentic Urban Planning AI Framework is implemented through a multi-agents collaboration framework that integrates the six logic parts (Analysis, Generation, Verification, Evaluation, Collaboration, Decision) operating across the three cognitive layers (Perception, Foundation, Reasoning). The framework supports two complementary collaboration methods tailored to different planning contexts:

\textbf{Method 1: Linear Individual Review}. In this workflow, individual planners sequentially review AI-generated planning proposals. Human planners independently rate proposals, provide comments, and suggest revisions. The AI system processes individual feedback to refine recommendations, enabling focused expert input without requiring group coordination. This method suits contexts where specialized expertise is needed (e.g., transportation engineers reviewing transit proposals) or when scheduling conflicts prevent simultaneous participation.

\textbf{Method 2: Group Discussion}. This workflow facilitates collective deliberation among multiple stakeholders. Planners engage in structured group discussions to evaluate AI recommendations collaboratively, surface conflicting priorities, negotiate trade-offs, and build consensus. The AI system captures group feedback, identifies areas of agreement and disagreement, and generates revised proposals that balance competing interests. This method is essential for contentious decisions requiring public input or multi-stakeholder negotiation (e.g., affordable housing siting, community facility allocation).

Both methods operate through a consistent \textit{Generation-Verification-Evaluation} pipeline anchored by the \textit{Human-AI Interface}. The multi-agents system generates planning alternatives, formally verifies regulatory compliance, and evaluates proposals against normative criteria. Human planners provide \textit{rating} (quantitative assessment of proposals), \textit{commenting} (qualitative feedback explaining concerns or suggestions), and request \textit{revision} (iterative refinement incorporating human input). This bidirectional interaction ensures that AI agents augment rather than automate planning decisions.

Critical to this multi-agents collaboration framework is the \textit{Analysis} component at the foundation, which synthesizes urban planning data from the Perception and Foundation layers to inform generation, and the \textit{Collaboration} component, which structures stakeholder input into machine-interpretable constraints and coordinates agent interactions. The final \textit{Decision} component integrates reasoning chains from multiple agents with human judgment, presenting \textit{Accept/Revise} options that preserve human agency in decision-making.

Algorithm \ref{alg:reasoning-pipeline} formalizes the core reasoning-verification-collaboration pipeline that operationalizes the Agentic Urban Planning AI Framework.

\begin{figure*}[!htbp]
\refstepcounter{algorithm}
\label{alg:reasoning-pipeline}
\begin{small}
\hrule
\vspace{2mm}
\noindent\textbf{Algorithm \thealgorithm:} Agentic Urban Planning CoT Pipeline with Human-AI Interface
\vspace{1mm}
\hrule
\vspace{2mm}
\begin{algorithmic}[1]
\STATE \textbf{Input:} Context $\mathcal{C} = \langle \mathcal{D}, \mathcal{K}, \mathcal{S} \rangle$, constraints $\mathcal{H}$, objectives $\mathcal{O}$, method $M$
\STATE \textbf{Output:} Proposal $p^*$ with complete reasoning chain $r^*$
\STATE
\STATE \textbf{// Phase 1: Analysis - Understand context and formulate strategies}
\STATE $\mathcal{H}, \mathcal{O} \leftarrow \text{HumanAI}.\text{InitRequirements}()$ \hfill // Collect human requirements
\STATE $features \leftarrow \text{Extract}(\mathcal{D})$ \hfill // Extract spatial, demographic, infrastructure features
\STATE $knowledge \leftarrow \text{RAG}.\text{RetrieveCases}(\mathcal{C}, \mathcal{K})$ \hfill // Query similar historical cases
\STATE $r_{ana} \leftarrow \text{Analyze}(features, knowledge)$ \hfill // CoT: diagnose issues, identify opportunities
\STATE $strategies \leftarrow \text{ProposeStrategies}(r_{ana}, \mathcal{H}, \mathcal{O})$ \hfill // Generate improvement strategies
\STATE \text{HumanAI}.\text{Present}($r_{ana}$, $strategies$) \hfill // Display analysis reasoning to user
\STATE
\STATE \textbf{// Phase 2: Generation - Create diverse proposals from strategies}
\STATE $\mathcal{P} \leftarrow \emptyset$, $\mathcal{R} \leftarrow \emptyset$ \hfill // Initialize proposal and reasoning sets
\STATE $regs \leftarrow \text{RAG}.\text{RetrieveRegs}(\mathcal{C})$ \hfill // Query relevant zoning codes and regulations
\FORALL{$s_j \in strategies$}
    \STATE $r_{gen} \leftarrow \text{ReasonStrategy}(s_j)$ \hfill // CoT: site selection, allocation logic
    \STATE $(p_j, r_{des}) \leftarrow \text{Design}(s_j, regs, \mathcal{H})$ \hfill // Design proposal following regulations
    \STATE $r_j \leftarrow r_{ana} \oplus r_{gen} \oplus r_{des}$ \hfill // Chain reasoning: analysis + generation + design
    \STATE $\mathcal{P} \leftarrow \mathcal{P} \cup \{p_j\}$, $\mathcal{R} \leftarrow \mathcal{R} \cup \{r_j\}$ \hfill // Add to candidate sets
    \STATE \text{HumanAI}.\text{Display}($p_j$, $r_j$) \hfill // Real-time visualization with reasoning
\ENDFOR
\STATE
\STATE \textbf{// Phase 3: Verification - Check constraint satisfaction symbolically}
\FORALL{$(p, r) \in \mathcal{P} \times \mathcal{R}$}
    \STATE $(valid, viols) \leftarrow \text{Check}(p, \mathcal{H})$ \hfill // Symbolic constraint checking (e.g., CSP solver)
    \STATE $r_{ver} \leftarrow \text{Explain}(viols)$ \hfill // Generate reasoning explaining verification result
    \IF{$\neg valid$}
        \STATE \text{HumanAI}.\text{LogViol}(viols); Remove $(p, r)$ \hfill // Log violations, filter invalid
    \ELSE
        \STATE $r \leftarrow r \oplus r_{ver}$ \hfill // Append verification reasoning to chain
    \ENDIF
\ENDFOR
\STATE
\STATE \textbf{// Phase 4: Evaluation - Assess impacts and score proposals}
\FORALL{$(p, r) \in \mathcal{P} \times \mathcal{R}$}
    \STATE $imp \leftarrow \text{AssessImpacts}(p, \mathcal{O})$ \hfill // Assess equity, sustainability, economic impacts
    \STATE $r_{eva} \leftarrow \text{ReasonImpacts}(imp)$ \hfill // CoT: explain impact assessment and trade-offs
    \STATE $score \leftarrow \text{Score}(imp, \mathcal{O})$ \hfill // Compute value alignment score (VAS)
    \STATE $r \leftarrow r \oplus r_{eva}$, store $(p, r, score)$ \hfill // Append evaluation reasoning, store
\ENDFOR
\STATE $ranked \leftarrow \text{Rank}(\mathcal{P}, scores)$ \hfill // Rank proposals by score
\STATE \text{HumanAI}.\text{DisplayRanked}($ranked$) \hfill // Show ranked alternatives to stakeholders
\STATE
\STATE \textbf{// Phase 5: Collaboration - Multi-role review and feedback collection}
\STATE $\mathcal{A} \leftarrow \{\text{Planner}, \text{Scientist}, \text{Citizen}, \text{Analyst}, \text{PolicyMaker}\}$ \hfill // Define stakeholder roles
\STATE $(p_{top}, r_{top}) \leftarrow \text{SelectTop}(\mathcal{P}, scores)$ \hfill // Select top-ranked proposal
\STATE \text{HumanAI}.\text{Present}($p_{top}$, $r_{top}$) \hfill // Present proposal with reasoning to stakeholders
\STATE $fb \leftarrow \text{CollectFeedback}(\mathcal{A}, p_{top}, M)$ \hfill // Method 1: Linear review; Method 2: Group discussion
\FORALL{$a \in \mathcal{A}$}
    \STATE $r_{rev}^a \leftarrow a.\text{Review}(p_{top}, r_{top})$ \hfill // Role-specific CoT (e.g., environmental concerns)
    \STATE $r_{top} \leftarrow r_{top} \oplus r_{rev}^a$ \hfill // Append all review reasoning to chain
\ENDFOR
\STATE
\STATE \textbf{// Phase 6: Decision - Synthesize feedback and finalize recommendation}
\STATE $conf \leftarrow \text{FindConflicts}(fb)$ \hfill // Identify conflicting opinions among stakeholders
\STATE $r_{con} \leftarrow \text{AnalyzeConf}(conf)$ \hfill // CoT: analyze nature and severity of conflicts
\IF{$conf \neq \emptyset$}
    \STATE $r_{res} \leftarrow \text{Resolve}(conf, fb)$ \hfill // CoT: explain conflict resolution strategy
    \STATE $(p_{rev}, r_{ref}) \leftarrow \text{Refine}(p_{top}, r_{res})$ \hfill // Refine proposal to address conflicts
    \STATE $r^* \leftarrow r_{top} \oplus r_{con} \oplus r_{res} \oplus r_{ref}$ \hfill // Complete reasoning chain
    \STATE \text{HumanAI}.\text{Present}($p_{rev}$, $r^*$) \hfill // Show final decision with full justification
    \STATE \textbf{return} $(p_{rev}, r^*)$ \hfill // Return revised proposal with reasoning
\ELSE
    \STATE $r_{dec} \leftarrow \text{Justify}(p_{top}, fb)$ \hfill // CoT: explain final decision rationale
    \STATE $r^* \leftarrow r_{top} \oplus r_{dec}$ \hfill // Complete reasoning chain with decision justification
    \STATE \text{HumanAI}.\text{Present}($p_{top}$, $r^*$) \hfill // Show final decision with full justification
    \STATE \textbf{return} $(p_{top}, r^*)$ \hfill // Return accepted proposal with reasoning
\ENDIF
\end{algorithmic}
\vspace{2mm}
\hrule
\end{small}
\end{figure*}

\section{Conclusion and Research Agenda}

This position paper has presented the Agentic Urban Planning AI Framework for reasoning-capable urban planning agents that integrates three cognitive layers (Perception, Foundation, Reasoning) with six logic components (Analysis, Generation, Verification, Evaluation, Collaboration, Decision) through a multi-agents collaboration framework. We have shown why urban planning decisions require explicit reasoning capabilities---multi-constraint satisfaction, transparent justification, normative deliberation---and demonstrated how this architecture addresses these requirements through value-based, rule-grounded, and explainable decision-making. 

\textbf{Open Research Challenges}. Realising this vision requires addressing five critical challenges:

\textit{(1) Constraint Knowledge Formalisation}: How do we encode urban planning knowledge---zoning codes, environmental regulations, design guidelines---in machine-interpretable form while maintaining flexibility? Key questions: formal languages for spatial/temporal/normative constraints; automatic extraction from regulatory documents; handling ambiguous or conflicting regulations.

\textit{(2) Reasoning Quality and Verification}: How do we ensure reasoning chains are correct and complete, especially when LLMs can generate plausible but invalid reasoning? Key questions: verifiers detecting constraint violations or logical errors; formal methods ensuring completeness; benchmarking reasoning quality when ground truth is normative.

\textit{(3) Scalability and Efficiency}: Real-world planning involves thousands of constraints and iterative refinement. Key questions: efficient search algorithms pruning invalid paths early; balancing reasoning depth with inference speed; caching/modular strategies enabling real-time interaction.

\textit{(4) Learning-Reasoning Integration}: What is the optimal division of labor between learning and reasoning components? Key questions: which tasks benefit most from learning (prediction) vs. reasoning (constraint satisfaction); how learned models provide probabilistic inputs to reasoning; detecting when learned estimates require verification.

\textit{(5) Fairness, Equity, and Value Alignment}: How do we ensure reasoning systems question historical biases and align with diverse stakeholder values? Key questions: explicit equity evaluation; transparent value elicitation and multi-stakeholder deliberation; auditing reasoning chains for hidden assumptions.

Beyond technical challenges, deployment raises policy questions about liability, transparency requirements, and democratic participation. Addressing these demands interdisciplinary collaboration among AI researchers, planners, legal scholars, and community stakeholders.

\textbf{The Path Forward}. The path forward requires collaboration between AI researchers, urban planners, and policymakers. Key priorities include: developing machine-readable planning knowledge bases (zoning codes, environmental standards); creating benchmarks for reasoning quality and constraint compliance; building agent architectures that integrate neuro-symbolic reasoning with test-time search; establishing verification and auditing protocols for AI planning assistants; and fostering interdisciplinary research on human-agent collaboration.

Urban planning decisions shape climate resilience, equity, opportunity, and health for generations. AI agents that reason transparently, verify constraints formally, and collaborate with human planners can help address unprecedented challenges---climate adaptation, housing crises, sustainable development---that demand both computational power and human wisdom. Reasoning is not merely beneficial---it is \textit{foundational} for AI systems that augment, rather than undermine, trustworthy planning.

The technical capabilities now exist. The challenge is to build systems worthy of the decisions they will help shape. The opportunity is immense. The time is now.

\appendix

\section{Evaluation Benchmark Metrics}

To assess reasoning-capable planning agents, we propose a comprehensive evaluation framework aligned with the three core requirements (value-based, rule-grounded, explainable) and six logic components of the Agentic Urban Planning AI Framework. Table \ref{tab:benchmark-metrics} presents benchmark metrics organized by evaluation dimensions.

\textbf{Formal Evaluation Metrics}. We define key metrics corresponding to the six logic components in Algorithm \ref{alg:reasoning-pipeline}:

\textit{Constraint Satisfaction Rate} (Verification): Measures the proportion of hard constraints satisfied by generated proposals.
\begin{equation}
\text{CSR}(p, \mathcal{H}) = \frac{|\{h \in \mathcal{H} : h(p) = \text{True}\}|}{|\mathcal{H}|}
\end{equation}

\textit{Reasoning Chain Quality} (Analysis, Generation): Assesses the logical coherence and completeness of reasoning chains.
\begin{equation}
Q(r) = \alpha \cdot \text{Coherence}(r) + \beta \cdot \text{Completeness}(r) + \gamma \cdot \text{Traceability}(r)
\end{equation}
where $\alpha + \beta + \gamma = 1$ are weighting parameters.

\textit{Value Alignment Score} (Evaluation): Quantifies alignment between AI decisions and normative planning principles.
\begin{equation}
\text{VAS}(p, \mathcal{O}) = \sum_{i=1}^{n} w_i \cdot \frac{o_i(p) - o_i^{min}}{o_i^{max} - o_i^{min}}
\end{equation}
where $o_i^{min}$ and $o_i^{max}$ define normalisation bounds for objective $o_i$.

\textit{Human-AI Collaboration Efficiency} (Collaboration): Measures interaction cycles required to reach acceptable proposals.
\begin{equation}
\text{HACE} = \frac{N_{accepted}}{N_{iterations} + \lambda \cdot T_{total}}
\end{equation}
where $N_{accepted}$ is the number of accepted proposals, $N_{iterations}$ is the total interaction cycles, $T_{total}$ is the total time, and $\lambda$ is a time penalty coefficient.

\textit{Decision Quality Score} (Decision): Quantifies the overall quality of final decisions considering multiple criteria.
\begin{equation}
\text{DQS}(p^*, r^*) = \omega_1 \cdot \text{CSR}(p^*, \mathcal{H}) + \omega_2 \cdot Q(r^*) + \omega_3 \cdot \text{VAS}(p^*, \mathcal{O})
\end{equation}
where $\omega_1 + \omega_2 + \omega_3 = 1$ are component weights.

\begin{table*}[!htbp]
\centering
\renewcommand{\arraystretch}{1.35}
\begin{small}
\setlength{\tabcolsep}{5pt}
\begin{tabular}{p{3cm}p{4.2cm}p{5cm}p{3.8cm}}
\hline
\textbf{Evaluation Dimension} & \textbf{Benchmark Metrics} & \textbf{Description} & \textbf{Pipeline Component} \\
\hline\hline
\multirow{2}{*}{\textbf{Analysis}} 
& Feature extraction accuracy & Precision/recall of spatial features and trade-off identification & Phase 1: Analysis \\
& Contextual understanding & Ability to interpret planning context $\mathcal{C}$ and identify key constraints & Phase 1: Analysis \\
\hline
\multirow{3}{*}{\textbf{Generation}} 
& Proposal diversity & Number of distinct valid alternatives generated per iteration & Phase 2: Generation \\
& Reasoning chain coherence & Human evaluation of CoT/ToT logical coherence (1-5 scale) & Phase 2: Generation \\
& Generation time & Time to produce $N_{samples}$ alternatives with reasoning (seconds) & Phase 2: Generation \\
\hline
\multirow{3}{*}{\textbf{Verification}} 
& Constraint satisfaction rate (CSR) & \% of hard constraints $\mathcal{H}$ satisfied by proposals (Eq. 1) & Phase 3: Verification \\
& Constraint violation rate & \% of proposals violating zoning, environmental, infrastructure constraints & Phase 3: Verification \\
& Verification latency & Time to verify proposal compliance (seconds per constraint) & Phase 3: Verification \\
\hline
\multirow{3}{*}{\textbf{Evaluation}} 
& Value alignment score (VAS) & Alignment with normative planning objectives $\mathcal{O}$ (Eq. 3) & Phase 4: Evaluation \\
& Equity impact assessment & Distributional fairness measures (Gini coefficient, accessibility gaps) & Phase 4: Evaluation \\
& Principle adherence & \% of decisions aligning with sustainability, equity principles & Phase 4: Evaluation \\
\hline
\multirow{3}{*}{\textbf{Collaboration}} 
& Collaboration efficiency (HACE) & Interaction cycles to reach acceptable proposals (Eq. 4) & Phase 5: Collaboration \\
& Feedback incorporation rate & \% of human critiques successfully integrated into revised proposals & Phase 5: Collaboration \\
& Stakeholder comprehension & Can non-experts understand AI rationales? (1-5 scale) & Phase 5: Collaboration \\
\hline
\multirow{3}{*}{\textbf{Decision}} 
& Decision quality score (DQS) & Overall quality combining CSR, Q, VAS (Eq. 5) & Phase 6: Decision \\
& Explanation completeness & \% of decisions with complete justifications citing sources and regulations & Phase 6: Decision \\
& Decision agreement & Agreement rate between AI recommendations and human planner decisions & Phase 6: Decision \\
\hline\hline
\end{tabular}
\end{small}
\caption{Benchmark metrics for evaluating reasoning-capable planning agents organised by the six logic components in Algorithm \ref{alg:reasoning-pipeline}. Each metric is mapped to its corresponding pipeline phase, ensuring alignment between formal evaluation (Equations 1-5) and operational implementation.}
\label{tab:benchmark-metrics}
\end{table*}

These metrics enable systematic evaluation of reasoning quality, constraint compliance, and human-AI collaboration effectiveness. Future research should develop standardized benchmark datasets for urban planning tasks with ground-truth constraint annotations, expert reasoning chains, and multi-stakeholder deliberation records.

\section*{Acknowledgments}

We thank our colleagues at the NUS Urban Analytics Lab for the discussions. This research is supported by NUS Research Scholarship (NUSGS-CDE DO IS AY22\&L GRSUR0600042). This research is part of the project Large-scale 3D Geospatial Data for Urban Analytics, which is supported by the National University of Singapore under the Start Up Grant R-295-000-171-133. This research is part of the project Multi-scale Digital Twins for the Urban Environment: From Heartbeats to Cities, which is supported by the Singapore Ministry of Education Academic Research Fund Tier 1.

\bibliography{aaai2026}


\end{document}